\documentclass[runningheads]{llncs}
\usepackage{graphicx}
\usepackage{cite}
\usepackage{times}
\usepackage{epsfig}
\usepackage{amsmath}
\usepackage{amssymb}
\usepackage{subcaption}
\usepackage{mwe}
\usepackage{acro}
\usepackage{amssymb}
\usepackage{xcolor,colortbl}
\usepackage{tabularx}
\usepackage{relsize}
\usepackage{pifont}
\usepackage{booktabs} 
\usepackage{multirow}
\usepackage{multicol}
\usepackage{adjustbox}
\usepackage{float}
\usepackage{tabularx}
\usepackage{array}
\DeclareMathSymbol{\shortminus}{\mathbin}{AMSa}{"39}

\usepackage[colorlinks,
            linkcolor=red,  
            anchorcolor=blue, 
            citecolor=green,   
            ]{hyperref}

\begin{document}
\newcommand{\etal}{\textit{et al}. }
\newcommand{\ie}{\textit{i}.\textit{e}., }
\newcommand{\eg}{\textit{e}.\textit{g}. }

\title{A Unified Learning Model for Estimating Fiber Orientation Distribution Functions on Heterogeneous Multi-shell Diffusion-weighted MRI}
%

\author{Tianyuan Yao\inst{1} \and
Nancy Newlin\inst{1} \and
Praitayini Kanakaraj\inst{1} \and
Vishwesh Nath\inst{3} \and
Leon Y Cai\inst{1} \and
Karthik Ramadass\inst{1} \and
Kurt Schilling\inst{2} \and
Bennett A. Landman\inst{1} \and
Yuankai Huo\inst{1}}
\colorlet{Mycolor1}{green!10!orange}
%
\institute{Vanderbilt University, Nashville TN 37215, USA \and
Vanderbilt University Medical Center, Nashville TN 37215, USA \and
NVIDIA Corporation, Santa Clara and Bethesda, USA\\
}
%
\maketitle              
\begin{abstract}
Diffusion-weighted (DW) MRI measures the direction and scale of the local diffusion process in every voxel through its spectrum in q-space, typically acquired in one or more shells. Recent developments in micro-structure imaging and multi-tissue decomposition have sparked renewed attention to the radial b-value dependence of the signal. Applications in tissue classification and micro-architecture estimation, therefore, require a signal representation that extends over the radial as well as angular domain. Multiple approaches have been proposed that can model the non-linear relationship between the DW-MRI signal and biological microstructure. In the past few years, many deep learning-based methods have been developed towards faster inference speed and higher inter-scan consistency compared with traditional model-based methods (e.g., multi-shell multi-tissue constrained spherical deconvolution). However, a multi-stage learning strategy is typically required since the learning process relies on various middle representations, such as simple harmonic oscillator reconstruction (SHORE) representation. In this work, we present a unified dynamic network with a single-stage spherical convolutional neural network, which allows efficient fiber orientation distribution function (fODF) estimation through heterogeneous multi-shell diffusion MRI sequences. We study the Human Connectome Project (HCP) young adults with test-retest scans. From the experimental results, the proposed single-stage method outperforms prior multi-stage approaches in repeated fODF estimation with shell dropoff and single-shell DW-MRI sequences.

\end{abstract}
\section{Introduction}

Diffusion-weighted magnetic resonance imaging (DW-MRI) is essential for the non-invasive reconstruction of the microstructure of the human \textit{in vivo} brain~\cite{basser1994estimation,van2012human,glasser2013minimal}. Substantial efforts have shown that other advanced approaches can recover more elaborate reconstruction of the microstructure \cite{MSMTCSDjeurissen2014multi, MAPMRIozarslan2013mean, HARDIdescoteaux1999high} and these methods are collectively referred to as high angular resolution diffusion imaging (HARDI). HARDI methods have been broadly proposed in two categories of single-shell acquisitions and multi-shell acquisitions (i.e., using multiple b-values). A majority of single-shell HARDI methods utilize spherical harmonics (SH) based modeling as in q-ball imaging (QBI) \cite{QBItuch2004q}, constrained spherical deconvolution (CSD) \cite{tournier2007robust}, and many others. However, SH-based modeling cannot directly leverage additional information provided by multi-shell acquisitions as the SH transformation does not allow for a representation of the radial complexity that is introduced by different b-value. SH has been combined with other bases to represent multi-shell data, e.g., solid harmonics \cite{SOLIDHARMONICSdescoteaux2011multiple}, simple harmonic oscillator reconstruction (SHORE) \cite{SHOREcheng2011theoretical}, and spherical polar Fourier imaging \cite{SPFIcheng2010model}.


Deep learning (DL) has revolutionized many different domains in medical imaging{~\cite{suzuki2017overview}, and DW-MRI parameter estimation is no different. Lots of DW-MRI methods have been developed that utilize the powerful data-driven capabilities of deep learning, yielding improved accuracy over conventional fitting when the acquisition scheme has a limited number of measurements~\cite{ muller2021rotation, xiang2023ddm}. However, most methods are only focused on the translation of single-shell data to DW-MRI parameters, and in contrast, the multi-shell methods get neglected due to the complexity associated with multi-shell data~\cite{nath2019inter, hansen2022contrastive}. Moreover, the SHORE-based DL methods typically used a multi-stage design~\cite{nath2019deep}. For instance, the algorithm must first optimize a specific optimal SHORE representation and then optimize the fiber orientation distribution function (fODF) estimation. Such methods are prone to overfitting, lower inference time, and complicated parameter tuning.



As shown in Fig~\ref{fig:problem}, in this paper, we propose a single-stage dynamic network with both the q-space and radial space signal based on a spherical convolutional neural network. We evaluated the resultant representation by targeting it to multi-shell multi-tissue CSD (MSMT-CSD). Both fiber orientation estimation and recovery of tissue volume fraction are evaluated. The contribution of this paper is three-fold:

$\bullet$ We proposed a unified dynamic network with the single-stage spherical convolutional neural network that can recover/predict microstructural measures.

$\bullet$ The proposed method is universally applicable to perform learning-based fODF estimation using a single deep model for various combinations of multiple shells.

$\bullet$ The proposed method achieved an overall superior performance compared with model-based and data-driven benchmarks.

\begin{figure}[t]
\begin{center}
\includegraphics[width=0.85\linewidth]{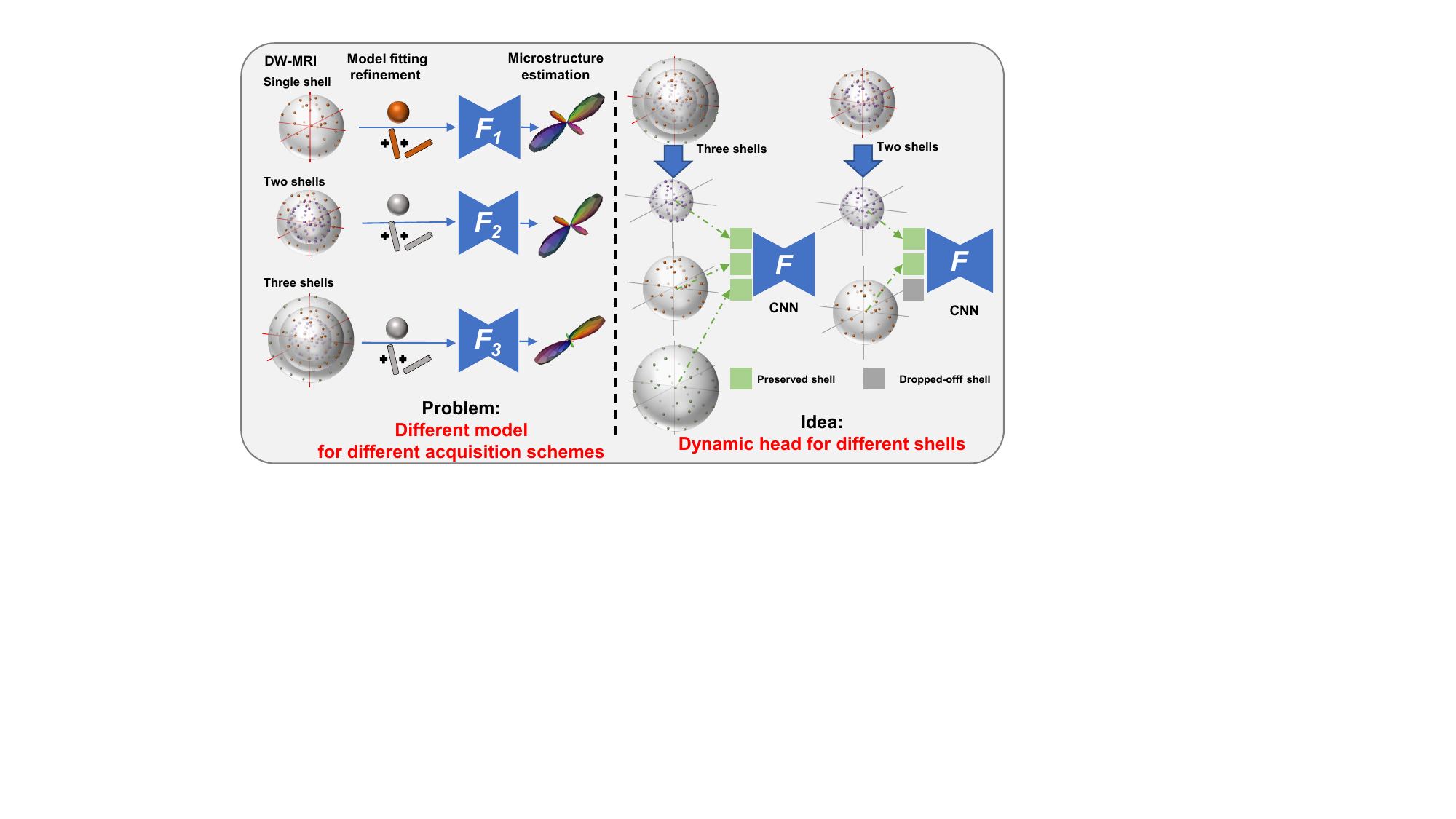}
\end{center}
   \caption{Utilizing multi-shell DW-MRI signals in deep learning usually requires independent models trained for each specific shell configuration as conventional SH-based modeling cannot directly leverage additional information (radial space) provided by multi-shell acquisitions. In our study, the dynamic head aims to improve the network expressiveness by learning and adaptively adjusting the first convolution layer for different shell configurations. }
\label{fig:problem}
\end{figure}

\section{Related work}
\subsection{Multi-Shell Multi-Tissue Constrained Spherical Deconvolution}

Multi-Shell Multi-Tissue Constrained Spherical Deconvolution (MSMT-CSD) \cite{MSMTCSDjeurissen2014multi} is a technique developed to overcome the limitations of traditional single-shell diffusion MRI methods, which are unable to resolve the complex fiber orientations of multiple tissue types in the brain. MSMT-CSD is able to separate the contribution of different tissue types (such as gray matter, white matter, and cerebrospinal fluid) to the diffusion signal by modeling the diffusion signal as a combination of multiple shells with different b-values. This modeling-based method has been a conventional method for multi-tissue micro-architecture estimation.

\subsection{Learning-based estimation}
Recently, machine learning (ML) and deep learning (DL) techniques have demonstrated their remarkable abilities in neuroimaging. Such approaches have been applied to the task of microstructure estimation \cite{nath2019deep}, aiming to directly learn the mapping between input DW-MRI scans and output fiber tractography~\cite{schilling2021fiber,cai2023convolutional} while maintaining the necessary biological characteristics and reproducibility for clinical translation. Such studies have illustrated that DL is a promising tool that uses nonlinear transforms to extract features from high-dimensional data. Data-driven approaches can be useful in validating the hypothesis of the existence of untapped information because they generalize toward the ground truth.


\section{Methods}
\subsection{Preliminaries}
Traditional deep learning frameworks are not generalizable to new acquisition schemes. This complicates the application of a DL model to data acquired from multiple sites. Our model aims to train a DL framework that can be adapted to an arbitrary number of available multi-shell DW-MRI sequences. To serve this motivation, we employ a dynamic head (DH) design to handle the multi-shell problem on the three most common b-values: 1000, 2000, and 3000 $s/mm^{2}$. Additionally, to tackle the problem of a varying number of gradient directions on each shell (b-value), we leverage the spherical CNNs with the traditional 'modeling then feeding to a fully connected network (FCN)' strategy. In this study, we employ the fODF estimation as our chosen task to perform assessments on different methods.
 
\subsection{Dynamic head design}
A dynamic head design in multi-modality deep learning~\cite{liu2022moddrop++} offers a flexible way to handle diverse data types within a single model, adapting its behavior to best suit the input modality. In our scenario, we intend to use a dynamic head that allows the neural network to effectively deal with diverse inputs from different shells by adapting its processing mechanism accordingly.

Note that with $K$ shells in our scheme, there are $2^K\shortminus1$ configurations. To improve the network expressiveness, we devise a dynamic head to adaptively generate model parameters conditioned on the availability of input shells. We use a binary code $m\in\mathbb{R}^K$ 
indicates that $m$ is a vector with $K$ real-valued entries or components. $K\in[0, 1]$ that 0/1 represent the absence/presence of each shell. To mitigate the large input variation caused by artificially zero-ed channels, we use the dynamic head to generate the parameters for the first convolutional layer.

\subsection{Spherical Convolution}
To extract features from DW-MRI signals, the first and most common deep learning network architecture applied to dMRI is the fully connected network(FCN)~\cite{aliotta2019highly, nath2019inter}, Conventionally these have been implemented the following: 

\begin{equation}
y = F_{FCN}(x |\theta_{FCN})
\end{equation}

Where $F_{FCN}$ is a fully-connected network with trainable parameters $\theta_{FCN}$, has signal input $x$ and $y$ is the ground truth dMRI parameters. Given a loss function, $L$ tailored for a specific downstream task and the function is learned by optimizing the trainable parameters $\theta_{FCN}$ and can be expressed as follows:

\begin{equation}\label{eq:learning_obj}
\tilde{\theta}_{FCN} = \arg\min_{\theta_{FCN}} L(yi, F(xi | \theta_{FCN})
\end{equation}

The dMRI signal $xi$ serves as the $i^{th}$ input for the network with corresponding ground truth output $yi$, and it does not consider the acquisition information, making the network unaware of the acquisition scheme. This lack of knowledge poses an issue when incorporating new data acquired at a different location with a distinct acquisition scheme. The accuracy of estimation from a new set of DW-MRIs depends on the consistency of the acquisition scheme with the training set. Additionally, the FCN's design does not account for rotational equivariance, which could result in requiring a varied range of tissue microstructure orientations in the training dataset for accurate estimation independent of fiber orientation.

Theoretically, Spherical CNNs offer an advantage over FCNs regarding both the robustness of the gradient scheme and the distribution of training data~\cite{ goodwin2022can,sedlar2021spherical}. The Spherical CNN's architecture differs from FCNs, but not in the conventional sense. Instead of convolution across multiple voxels, Spherical CNNs perform convolution over the spherical image space. Hence, like FCNs, they are voxelwise networks. At each voxel, the spherical image is created from the dMRI signals and their corresponding gradient scheme. This architecture can naturally address the limitations of FCNs in two ways. Firstly, unlike FCNs, Spherical CNNs inherently recognize the gradient scheme present in their input, as illustrated by the following equation: 

\begin{equation}
y = F_{S-CNN}(x, G | \theta_{S-CNN}) 
\end{equation}

Here, $F_{S-CNN}$ represents the Spherical CNNs, characterized by their trainable parameters $\theta_{S-CNN}$. One of the distinct advantages of Spherical CNNs is their explicit consideration of the gradient scheme in the input. This capability enables them to adeptly manage variations in gradient schemes that may arise from different acquisition protocols or disparate imaging sites. Moreover, owing to their inherent spherical structure, Spherical CNNs can more effectively handle the distribution of training data that resides in a spherical domain. Cumulatively, the unique attributes of spherical convolution present significant improvements in the accuracy and robustness when analyzing diffusion MRI signals. During the training phase, the shared network $F_{S-CNN}$ utilizes the input data. This data comprises $2^K\shortminus1$ distinct shell configurations, which can be described as 

\begin{equation}
\tilde{x}^{k} = \delta^{k} x^{k},  (k \in {1,...,K})
\end{equation}

Where $\delta^{k}$ is a Bernoulli selector variable that can take on values in {0, 1}. By aiming at diverse diffusion properties denoted by $y$ and combined with the dynamic head setting, the learning objective at the $i^{th}$ input can be articulated as:

\begin{equation}\label{eq:learning_obj}
\tilde{\theta}_{S-CNN} = \arg\min_{\theta_{S-CNN}}L(yi, F(\tilde{x}^{k}i ,Gi | \theta_{S-CNN}))
\end{equation}

\section{Experiments}

\subsection{Data and Implementation Details}
We have chosen DW-MRI from the Human Connectome Project - Young Adult (HCP-ya) dataset~\cite{van2013wu, glasser2013minimal}, 45 subjects with the scan-rescan acquisition were used (a total of 90 images). The acquisitions had b-values of 1000, 2000, 3000 $s/mm^{2}$ with 90 gradient directions on each shell. A T1 volume of the same subject was used for WM segmentation using SLANT\cite{SLANT}. All HCP-ya DW-MRI was distortion corrected with top-up and eddy\cite{jenkinson2012fsl}. 30 subjects are used as training data while 10 subjects were used as evaluation and 5 subjects as testing data.

We performed shell extraction on all the data. Every subject has seven different shell configurations which are the permutations of all three b-values $\{$$\{1K, 2K, 3K\}$, $\{1K, 2K\}$, $\{2K, 3K\}$, $\{1K, 3K\}$, $\{1K\}$, $\{2K\}$, $\{3K\}$$\}$. Ground-truth fODF maps were computed from MSMT-CSD using the DIPY library with the default settings ~\cite{DIPY}. $8^{th}$ order SH were chosen for data representation with the 'tournier07' basis~\cite{tournier2007robust}. The white matter fODF and the volume fraction which refers to the proportion of the volume of the voxel that is occupied by each tissue type, are combined together as the targeted sequence. 

Inspired by Nath et al.\cite{nath2019deep}, we employed the simple harmonic oscillator-based reconstruction and estimation (SHORE) as another baseline representation. SHORE modeling is known to capture the complex diffusion signal across different b-values without resorting to multi-compartment models, where the SHORE basis function is given by $Z_{nlm}(q, \Theta) = R_{n}(q) Y_{lm}(\Theta)$. As for the single shell dMRI signal, $|q|$ is constant, and the variability in $E(\mathbf{q})$ is primarily captured by $Y_{lm}(\Theta)$. The richness of the model (i.e., maximum order N) is likely needed to be limited for single-shell data to avoid overfitting. Thus, the $6^{th}$ radial order SHORE basis is employed as a baseline representation for both single-shell and multi-shell dMRI signals to fit the fiber ODF. The SHORE scaling factor $\zeta$ defined in units of $ mm^{-2}$ as $\zeta = 1/8\pi^{2}\tau MD$ is calculated based on the mean diffusivity (MD) obtained from the data. Given that both SHORE base signal ODF and SH base fiber ODF have the same underlying information. We apply deep learning to map the intricate relationships and patterns from one representation to another.

\subsection{Experimental setting}
We first trained separate models for each shell configuration. The models consist of four fully connected layers with ReLU activation function. The number of neurons per layer is 400, 48, 200, and 48. The input is the $1 \times 50$ vector of the shore basis signal ODF, and the output is the combination $1 \times 45$ vectors of the SH basis WM fODF and the $1 \times 3$ vector of tissues fraction. The models are then tested on the different shell configurations. By simply feeding all the shell configuration data (all labeled with reconstructed fODF from data with all shells) to the FCN as a baseline 'unified' deep learning model.

We assess the impact of dynamic head strategy by evaluating the performance of the unified models against independent models trained for each specific shell configuration. Furthermore, the generalizability of the different representations with dynamic head designs was assessed. For the spherical convolution, we used an architecture known as the hybrid spherical CNN as described in ~\cite{cobb2020efficient}. The architecture consists of a $S^2$ convolutional layer and a $SO(3)$ convolutional layer and is followed by three channel-wise activations and two restricted generalized convolutions until the final restricted generalized convolution maps down to a rotationally invariant representation. The specific network parameters follow the spherical MNIST experiment~\cite{cobb2020efficient}. The diffusion signals from different shells are 1-to-1 densely sampled to map between six directional dMRI signals and the 6 independent values of the diffusion tensor. After the rotational invariant features are extracted. They are concatenated and fed into fully connected layers(the same hidden size as above) which perform the final estimation.

\subsection{Evaluation metric}
To evaluate the predictions from the proposed methods, we calculated the mean squared error of the volume fractions with ground truth sequences. Then we compute the angular correlation coefficient (ACC, Eq.~\ref{ACC}) between the predicted fODF and the ground truth fODF over the white matter region. ACC is a generalized measure for all fiber population scenarios. It assesses the correlation of all directions over a spherical harmonic expansion. In brief, it provides the estimate of how closely a pair of fODFs are related on a scale of -1 to 1, where 1 is the best measure. Here ‘u’ and ‘v’ represent sets of SH coefficients.

\begin{equation}\label{ACC}
\begin{split}
ACC= \frac{\sum_{k=1}^{L}\sum_{m=-k}^{k}(u_{km})(v^*_{km})}{[\sum_{k=1}^{L}\sum_{m=-k}^{k}|u_{km}|^2]^{0.5}\cdot[\sum_{k=1}^{L}\sum_{m=-k}^{k}|v_{km}|^2]^{0.5}} 
\end{split}
\end{equation}

\section{Experimental Results}
We compared the performances of the unified models against independent models trained for each specific shell configuration. A qualitative result of fODF predictions and GT are shown in Fig~\ref{fig: vis}. As shown in Table~\ref{table:model1}, the independent models that are trained are thus more likely to outperform others in their own shell configuration and these models can be considered as the upper bounds for each shell configuration. With the dynamic head settings, the unified model with spherical convolution outperforms the other models in the single shell configuration. Additionally, the ACC is a sensitive generalized metric, the performances need further evaluation. We assessed how good our prediction was by evaluating the scan/rescan consistency and volume fraction prediction~\ref{table:model2}.

\begin{figure}[t]
\begin{center}
\includegraphics[width=0.85\linewidth]{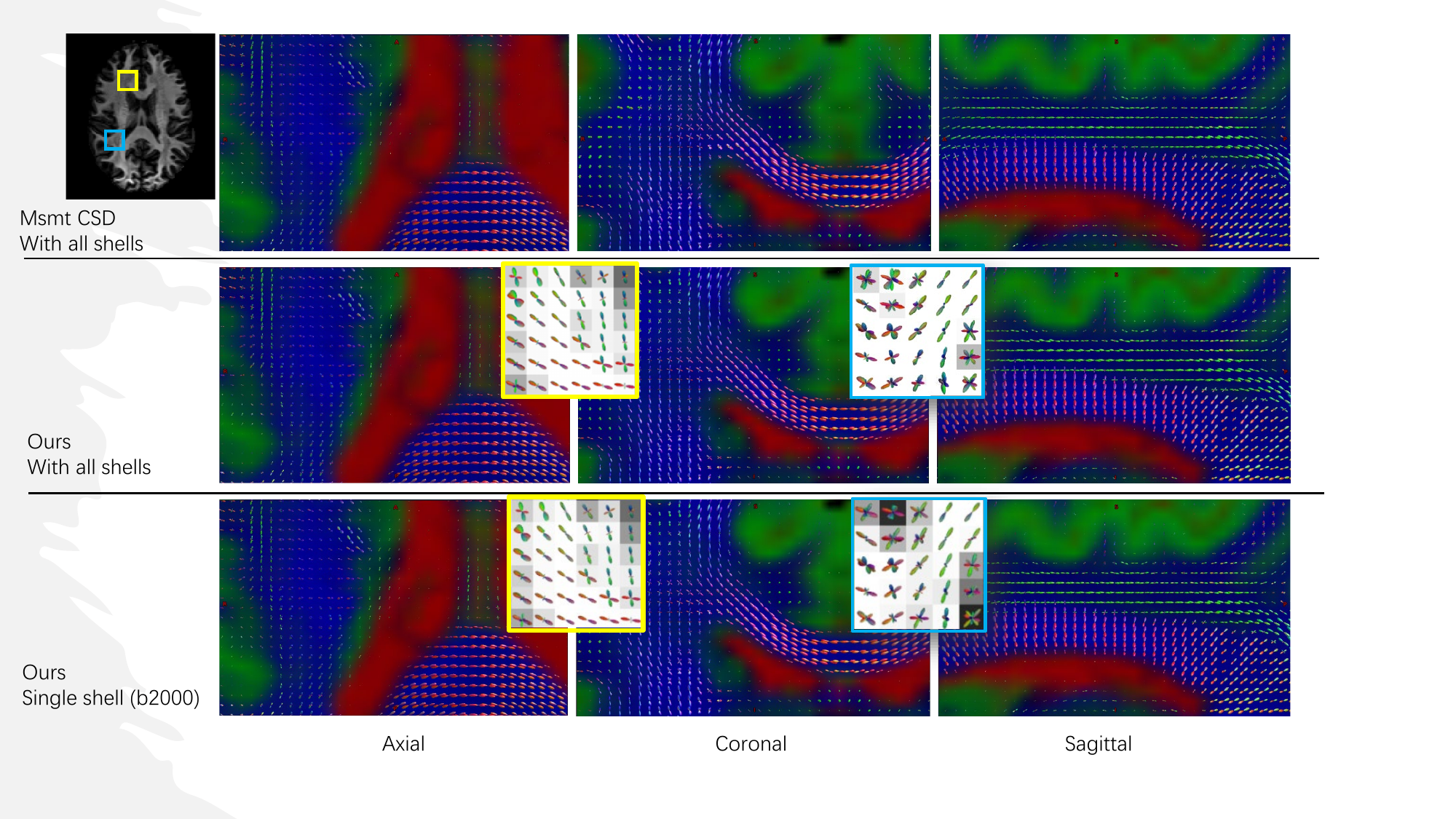}
\end{center}
   \caption{This is a visualization of the fODF prediction and the correlation with the GT in different views. The background of the zoom-in patches shows the ACC spatial map with the GT signals.}
\label{fig: vis}
\end{figure}

\newcolumntype{P}[1]{>{\centering\arraybackslash}p{#1}}
\begin{table}[htbp]
\caption{Performances of the unified models against independent models in different shell configurations}
\centering
\small
\begin{tabular}{|P{2.6cm}|P{1cm}|P{1cm}|P{1cm}|P{1cm}|P{1cm}|P{1cm}|P{1cm}|P{1.4cm}|}
 \hline
 \textbf{Model} & \textbf{$C_{1}$} & \textbf{$C_{2}$} & \textbf{$C_{3}$} & \textbf{$C_{1,2}$} & \textbf{$C_{2,3}$} & \textbf{$C_{1,3}$} & \textbf{$C_{1,2,3}$} & \textbf{Ave.} \\ \hline 
 $M_{1}$ & \textcolor{blue}{0.808} & 0.725 & 0.732 & 0.752 & 0.734 & 0.751 & 0.788 & 0.756 \\
$M_{2}$ & 0.762 & \textcolor{blue}{0.815} & 0.756 & 0.744 & 0.749 & 0.745 & 0.774 & 0.764 \\
$M_{3}$ & 0.757 & 0.724 & \textcolor{blue}{0.814} & 0.734 & 0.753 & 0.760 & 0.779 & 0.760 \\
$M_{1,2}$ & 0.745 & 0.732 & 0.743 & \textcolor{red}{ 0.831} & 0.788 & 0.778 & 0.789 & 0.772 \\
$M_{2,3}$ & 0.734 & 0.744 & 0.738 & 0.802 & \textcolor{blue}{0.825} & 0.786 & 0.786 & 0.774 \\
$M_{1,3}$ & 0.737 & 0.745 & 0.745 & 0.785 & 0.793 & \textcolor{red}{ 0.832} & 0.784 & 0.774 \\
$M_{1,2,3}$ & 0.752 & 0.734 & 0.742 & 0.762 & 0.756 & 0.772 & \textcolor{red}{ 0.853} & 0.767 \\ \hline
All Data Feeding & 0.789 & 0.793 & 0.794 & 0.801 & 0.799 & 0.803 & 0.814 & 0.799 \\
DH w. SHORE~\cite{SHOREcheng2011theoretical} & 0.782 & 0.788 & 0.784 & 0.823 & 0.817 & \textcolor{blue}{0.825} & \textcolor{blue}{0.843} & 0.809 \\ DH w. SH & 0.805 & 0.809 & \textcolor{blue}{0.814} & 0.818 & 0.812 & 0.812 & 0.832 & \textcolor{blue}{0.815} \\
DH w. SC (Ours) & \textcolor{red}{ 0.816} & \textcolor{red}{ 0.82} & \textcolor{red}{ 0.816} & \textcolor{blue}{0.827} & \textcolor{red}{ 0.828} & 0.824 & 0.837 & \textcolor{red}{ 0.824} \\ \hline
\end{tabular}
    \caption*{Table1: $M_{i}$, where i $\in [1,2,3]$ indicates the model is only trained on that shell configuration. $C_{i}$ indicates the testing data in that shell configuration. The best and second best performances are denoted by the \textcolor{red}{red} mark and \textcolor{blue}{blue} mark. The average metrics of ACC are listed in the last column }
\label{table:model1}
\end{table}

\begin{table}[htbp]
\caption{FODF prediction assessment}
\centering
\small
\begin{tabular}{|P{2.5cm}|P{3cm}|P{2.5cm}|P{3.5cm}|}
 \hline
 \textbf{Model} & \textbf{Shell configuration} & \textbf{Tissue proportion prediction} & \textbf{Scan-rescan consistency}  \\ \hline 
 \multirow{6}{*}{Single model} & 1K & 8.45E-04 & 0.862 \\
  & 2K & 7.92E-04 & 0.865 \\
 & 3K & 8.63E-04 &  0.857 \\
 & 1K, 2K & 7.32E-04 & 0.856 \\
 & 2K, 3K & 7.49E-04 & 0.86 \\
 & 1K, 3K & 8.02E-04 & 0.862 \\
 & 1K, 2K, 3K & 6.38E-04 & 0.865 \\   \hline
\multirow{6}{*}{DH w. SC}  & 1K & 7.27E-04 & 0.855 \\
  & 2K & 7.12E-04 & 0.86 \\ 
  & 3K & 7.35E-04 & 0.858  \\
  & 1K, 2K & 7.01E-04 & 0.858 \\ 
 & 2K, 3K & 6.79E-04 & 0.86 \\ 
 & 1K, 3K & 6.82E-04 & 0.864 \\
 & 1K, 2K, 3K & 5.92E-04 & 0.861 \\    \hline
\multicolumn{2}{|c|}{Silver standard : MSMT-CSD~\cite{MSMTCSDjeurissen2014multi}} &  & \textcolor{red}{0.856} \\ \hline
\end{tabular}
\label{table:model2}
\caption*{Table2: Reconstruction results from msmt-CSD are applied as silver standard in the evaluation. Wilcoxon signed-rank test is applied as a statistical assessment for scan-rescan consistency evaluation. It has a significant difference ($p < 0.001$) compared with WM fODF. The MSE is reported for evaluation of VF predictions. The ACC between scan/rescan DW-MRI over WM regions is reported.}
\end{table}

\section{Conclusion}
In this paper, we propose a single-stage dynamic network with both the q-space and radial space signal based on a spherical convolutional neural network. Integrating dynamic head and spherical convolution removes the need to retrain a new network for a known b-value of DW-MRI. Besides, adjusting the last multi-layer regression network to different targets, this plug-and-play design of our method is potentially applicable to a wider range of diffusion properties in neuroimaging.



%
%
\bibliographystyle{splncs04}
\bibliography{main}
%




\end{document}